\documentclass[journal]{IEEEtran}
\ifCLASSINFOpdf
\else
\fi

\usepackage{graphicx}
\usepackage{booktabs,caption}
\usepackage{dblfloatfix}
\usepackage{booktabs}
\usepackage{multirow}
\usepackage{makecell}

\usepackage{tabularx,colortbl}
\usepackage{subcaption}

\usepackage{soul,color}
\soulregister\cite7
\soulregister\ref7
\soulregister\pageref7

\usepackage[most]{tcolorbox}
\newtcolorbox{highlighted}{colback=yellow,coltext=red,breakable}
\usepackage{lipsum}

\begin{document}
%
\title{Domain Generalization in Biosignal Classification.}
%
%
%

\author{Theekshana~Dissanayake,~Tharindu~Fernando,~\IEEEmembership{Member,~IEEE,}~Simon~Denman,~\IEEEmembership{Member,~IEEE},~Houman Ghaemmaghami,~Sridha~Sridharan~\IEEEmembership{Life Senior Member,~IEEE,}~and~Clinton~Fookes,~\IEEEmembership{Senior Member,~IEEE}
\thanks{
Theekshana Dissanayake, Tharindu Fernando, Simon Denman, Sridha Sridharan and Clinton Fookes are from the SAIVT Research group at the Queensland University of Technology, Australia}}

%
%

\markboth{Journal of \LaTeX\ Class Files,~Vol.~14, No.~8, August~2015}%
{Shell \MakeLowercase{\textit{et al.}}: Bare Demo of IEEEtran.cls for IEEE Journals}
%



\maketitle

\begin{abstract}

\textit{Objective:} When training machine learning models, we often assume that the training data and evaluation data are sampled from the same distribution. However, this assumption is violated when the model is evaluated on another unseen but similar database, even if that database contains the same classes. This problem is caused by domain-shift and can be solved using two approaches: domain adaptation and domain generalization. Simply, domain adaptation methods can access data from unseen domains during training; whereas in domain generalization, the unseen data is not available during training. Hence, domain generalization concerns models that perform well on inaccessible, domain-shifted data. \textit{Method:} Our proposed domain generalization method represents an unseen domain using a set of known basis domains, afterwhich we classify the unseen domain using classifier fusion. To demonstrate our system, we employ a collection of heart sound databases that contain normal and abnormal sounds (classes). \textit{Results}: Our proposed classifier fusion method achieves accuracy gains of up to 16\% for four completely unseen domains.  \textit{Conclusion}: Recognizing the complexity induced by the inherent temporal nature of biosignal data, the two-stage method proposed in this study is able to effectively simplify the whole process of domain generalization while demonstrating good results on unseen domains and the adopted basis domains.  \textit{Significance}: To our best knowledge, this is the first study that investigates domain generalization for biosignal data. Our proposed learning strategy can be used to effectively learn domain-relevant features while being aware of the class differences in the data.
\end{abstract}


\IEEEpeerreviewmaketitle

\section{Introduction}
Due to a lack of generalization capability, traditional machine learning models tend to perform poorly on data from completely unseen domains. In the machine learning community, this phenomenon is termed the domain-shift problem, and it significantly affects the usability of classifiers for real-world applications. 

The domain-shift problem can be solved using two techniques: domain adaptation and domain generalization. Simply, if some data in the new target domain is available at the training stage, we can use domain adaptation to transform the trained model into the new domain. However, in real-world scenarios, we can't access the unseen data in the new target domain at the training stage. Hence, domain generalization techniques deal with designing models that perform well on inaccessible domain-shifted data. Therefore, domain generalization can be seen as a more complex task compared to domain adaptation \cite{Rahman2020Correlation-awareGeneralization}. Examining the literature, we find that researchers in the medical field have been able to develop successful techniques for domain adaptation \cite{Nidadavolu2019Low-ResourceCycle-Gans,Hou2019PredictionAdaptation,He2020TransferApproach}; however there has not been any significant effort in applying  domain generalization in the medical field, and the problem remains open and requires further attention.

In the medical domain, the generalization ability of the model is important for two reasons. Firstly, a false prediction made by a model may have a serious consequences for the patient. Secondly, since medical practices and protocols are continuously reviewed and updated, a machine learning model trained on past data may perform poorly on newly collected data captured using different methods or devices \cite{pubmed}. 

Even though there is an essential need for designing generalized models for clinical applications, research effort in the medical domain lags in exploring these methods compared to other machine learning application fields (eg: object recognition, robotic navigation). While other fields have conducted such investigations, many of the techniques developed in those fields can not be directly applied to medical modalities such as biosignals, because of their inherent temporal nature. Also, a majority of those studies have used synthesized data that may not correctly represent natural domain-shifts. In the medical domain we have the advantage of being able to acquire real domain-shifted data, that is collected by and labelled by experts. 

Addressing  this essential need, we propose a two stage deep learning framework to achieve domain generalization on medical signals. The first stage of the framework consists of a domain recognition deep learning model while the second stage implements a classifier ensemble to effectively classify the given unseen domains or instances. To demonstrate our method, we use the PhysioNet~\cite{Goldberger2000PhysioBankSignals.} heart sound database and four other similar heart sound databases. In this setting, the final output of our classifier-framework will indicate whether a given signal is normal or abnormal. The following list highlights the main contributions of our study:

\begin{enumerate}

\item To our knowledge, this is the first investigation that aims to solve the domain generalization problem in the context of biosignal classification tasks in the machine learning domain. 

\item We extend the work of Schroff and Philbin \cite{Schroff2015FaceNet:Clustering} on Triplet loss to design a deep learning model that can learn domain-relevant embeddings. The model demonstrate uniform prediction capability across multiple unseen domains, achieving classification accuracy gains of up to 8\% before fusion (average: 2.65\%).

\item We also evaluate the use of Bayesian models for domain recognition task, and demonstrate how the uncertainty measured by such methods can enable domain generalisation.

\item Our framework achieves accuracy gains of up to 16\%~(average: 7.69\%) for four different and diverse, unseen domains compared to the baseline method proposed by \cite{Mancini2018BestNets}.

\end{enumerate}

The rest of the paper is organized as follows. Section \ref{related} addresses  related work and explains the research gap that currently prevails in applying domain generalisation in the medical domain. Section \ref{method} presents the methodology and describes how we solve this essential problem by employing deep learning. In Section \ref{results} we present our findings and visualize how new domains are being distributed among the basis domains (or known domains). Finally, Section \ref{discuss} provides an summary of our research, and some of the difficulties we encountered.

\section{Related Work\label{related}}

As noted, the medical domain does not have prior studies on domain generalization. Therefore, in this section, we discuss some domain generalization studies from other related machine learning application fields, and highlight how effectively those studies can be applied to solve domain generalization for biosignal data. 

According to Li~et~al.~\cite{Li2018LearningGeneralization}, existing methods for domain generalization can be categorized into three classes. The first is based on representing a new domain using an existing set of basis domains, and then classifying the new domain using classifier fusion. The second approach deals with finding an underlying globally-shared domain that represents the hidden patterns in all known domains. Finally, the third method deals with discovering domain-invariant features for training a classifier. 

In their investigation, Mancini~et~al.~\cite{Mancini2018BestNets} propose a deep learning framework that uses domain-relevant weighted fusion to classify an unseen instance from an unseen domain. They have used a joint-learning approach to train the domain recognizer and domain-specific classifiers on image data. A similar variant of the proposed system applied to place recognition for robotic vision can be found in \cite{Mancini2018RobustGeneralization}. Examining other methods that use classifier fusion, the recent study by Niu~et~al.~\cite{Niu2018AnRecognition} introduces an exemplar SVM-based multi-classifier system for domain generalization. Their work is an extension of the study performed by Xu~et~al.~\cite{Xu2014ExploitingGeneralization}, which proves that combining such classifiers produces more generalized models. 

In their paper, Li~et~al.~\cite{Li2017DeeperGeneralization} introduce a generalised convolution neural network for image classification based on finding a globaly-shared domain. In their study, they extend the work of Khosla~et~al.~\cite{Khosla2012UndoingBias} on domain generalization in linear models to the deep neural networks. Khosla et al.~\cite{Khosla2012UndoingBias}
assumes the domain-shift will have the mathematical form of $\Theta_i = \Theta_0 + \Delta_i$, where $\Theta_0$ represents the parameter of the universal domain and $\Delta_i$ represents the parameter shift for the $i^{th}$ domain ($\Theta_i$). 

Considering domain-invariant feature learning methods, Li~et~al.~\cite{Li2018DomainLearning} proposes an adversarial learning technique to design a deep learning model that can generate a feature space with two properties: it possess domain-invariant features and those features are capable of discriminating classes. To evaluate the performance of their method, they have used the rotated-MNIST image database \cite{Larochelle2007AnVariation}. In another investigation, Li~et~al.~\cite{Li2018DeepNetworks} suggests an end-to-end conditional invariant deep domain generalization approach to learn domain-invariant representations. They have also evaluated their models on the rotated-MNIST database and PACS dataset \cite{Li2017DeeperGeneralization,Larochelle2007AnVariation}. Their results suggest that the proposed learning protocol leads to a better representation of domain-invariant features. 

Reflecting on studies in the literature, we argue that biosignal classification-based domain generalization can be seen as a complex task due to the inherent temporal nature of the data. Furthermore, factors such as evolving medical practices, performance of acquisition devices and the experiment settings employed add additional real variations to the collected data, compared to synthesized data used in other discussed studies \cite{Li2018DomainLearning,Li2017DeeperGeneralization}.  Recognizing these limitations, domain-shifts in biosignal data will provide an opportunity to design deep learning models that operate on real data, which ultimately provide valuable insights into developing robust domain generalization methods.

In our investigation, we explore how to achieve domain generalization in biosignal-based classification tasks. For this, we adopt the domain basis technique with a two-stage learning process (i.e domain recognizer and classifier fusion). By dividing the problem into two stages, we believe that the classifiers will be able to individually learn complex patterns in the data related to domain-shifts and classes. Also, more importantly, those models will be able to perform associatively to solve the final problem. Being the first investigation on domain generalization in medical signals, our investigation introduces this crucial problem to the community, and we believe insights gained from our study will help to advance the state of domain generalization methods in this and  other fields. 

\section{Methodology\label{method}}

As discussed, in this study we investigate how to use deep learning models to achieve domain generalization in biosignal classification tasks. The proposed framework contains two components: the domain recognizer and the classifier ensemble. The task of the domain recognizer is to identify a set of the most suitable domains for a given unseen (i.e new) instance, while the classifier ensemble is effectively identifying the class of the given instance using the insights provided by the domain recognizer. We use the term domain to refer to a database that possesses instances from similar classes extracted using different protocols, devices, etc. 

In the context of domain generalization, if a particular domain in a set of known domains has a unique signature which is universal for all instances within that particular domain, and an unseen instance sampled from a particular unseen domain possesses that same signature, then we say that the unseen instance belongs to that particular domain. Here, we define this association as the domain-relevance. It should be noted that this relationship is complex, which we are trying to determine using deep learning. Regarding this analogy, we can identify the following relationships:

\begin{table*}[b!]
    \centering
    \begin{tabular}{|p{3.0cm}|p{0.7cm}|p{0.7cm}|p{2.2cm}|p{2.8cm}|p{1.0cm}|p{3.7cm}|}
    \hline
    Database &  N & AN & Environment & Pathology Annotations &  Patients & Device Specifications \\ \hline
    
    PhysioNet(a)~\cite{Goldberger2000PhysioBankSignals.} & 21470 & 45502 & In-house+Hospital  & MVP, Aortic and Benign & 121 & Welch Allyn Meditron \\
    
    PhysioNet(b)~\cite{Goldberger2000PhysioBankSignals.} & 6065 & 2085 & Hospital & Murmur sounds  & 106 & 3M Littmann E4000 \\ 
    
    PhysioNet(c)~\cite{Goldberger2000PhysioBankSignals.} & 1583 & 6609 & Hospital & MR, AS & 31 & AUDIOSCOPE \\
    
    PhysioNet(d)~\cite{Goldberger2000PhysioBankSignals.} & 965 & 2075 & Hospital+Lab & None & 38 & Infral Corp. Prototype  \\
    
    PhysioNet(e)~\cite{Goldberger2000PhysioBankSignals.} & 181254 & 18606 & Hospital+Lab & Murmurs & 509 & 3M Littmann, MLT201/Piezo \\ 
    
    PhysioNet(f)~\cite{Goldberger2000PhysioBankSignals.} & 11207 & 4649 & Hospital & None & 112 & JABES Electronic stethoscope \\
    
    FetalDB~Mother's \cite{Samieinasab2015FetalSeparation} & 5479 & 392 & Lab & Preeclampsia, Mild MP & 109 & JABES Electronic stethoscope \\
    
    FetalDB~Child's\cite{Samieinasab2015FetalSeparation} & 26995 & 340 & Lab & High fetal heart rate  &  119 & JABES Electronic stethoscope \\
    
    M3dicine-Human~\cite{Fernando2019HeartAttention} & 8287 & 3407 & Hospital+Lab & None & 158 & Stethee digital stethoscope \\
    
    M3dicine-Animal~\cite{Fernando2019HeartAttention} & 14593 & 13181 & Hospital+Lab & None & 376 & Stethee digital stethoscope \\
    \hline
    \end{tabular}
    \caption{Databases used in the analysis. \textbf{N}: Normal instances count, \textbf{AN}: Abnormal instance count (1s windowed signals)}
    \label{dbsall}
\end{table*}

\begin{enumerate}
    \item One Instance to One Domain \textbf{(One-to-One)}: One instance can be related to only one known domain. Here, we are not considering the class of the given instance, the relationship (or the signature) we are trying to identify is domain-relevant.
    
    \item One Instance to Many Domains \textbf{(One-to-N)}: Given that our domains are diverse, one instance can fall into multiple domains with different degrees of associations.  
    
    \item One Instance to None \textbf{(One-to-None)}: This scenario can arise if the number of adopted known domains is less or the data within selected domains are not sufficiently diverse. Occurrences of this type may also arise if the given instance is a noisy output from the sensor system or an outlier.
\end{enumerate}

Considering these categories, if all unseen domain instances represent the \textbf{One-to-One} relationship, then ultimately, the entire unseen domain will be related to a single known domain. However, we believe the \textbf{One-to-N} relationship can be seen as the typical outcome in the biomedical domain, as most databases found in this field for a particular task are a collection of sub-databases. For instance, the PhysioNet database for heart sound analysis is a collection of six databases (see Table \ref{dbsall}). Even though some studies have treated the combined database as a single domain, we believe this is an oversimplification of the actual conditions and it limits the adaptation to the new domain. Hence, in our experiments we will investigate the advantage of domain separation, and how it impacts the generalization ability of the classifier. 

Equation \ref{probdef} represents the problem definition of the domain recognizer where, given a new instance ($I$), the deep learning model ($f$) should be able to determine the relationship factor ($\beta$) between the given instance and the known $n$ domains ($D_i$). 
\begin{equation}
\label{probdef}
    f(I) = \beta_{i}\times D_{i},~~\{i \in [1,2,\dots,n],~I \in D_{us}\}. 
\end{equation}
Then, the entire new domain (or unseen $D_{us}$) should be classified using an ensemble of classifiers. To determine $f$, we use two different learning strategies. The first method is inspired by recent advancements in One-Shot Learning \cite{Schroff2015FaceNet:Clustering}, and the second method is based on Bayesian Networks. To the best of our knowledge this is the first work investigating the domain adaptation task with bio-signals using such learning strategies. As such we investigate the merits of each of the classification techniques in order to determine the best framework for the task at hand.

After recognizing the instance-domain relationship, we use an ensemble of deep learning models to classify a new instance. In this stage, we emphasize that if the given instance belongs to a particular domain or a set of domains, then, a model trained on that domain should be able to classify that instance accurately. Furthermore, the fusion process in the classifier ensemble should be able to mitigate false classification of the given instance if it belongs to multiple domains. This process can be expressed by Equation \ref{probinf} where $g$ denotes the fusion function which takes the instance-domain relationship ($[\beta_i]_n$) and the ensemble of classifiers ($[clsf_i]_n$), and outputs the relevant class.   
\begin{equation}
\label{probinf}
    g([\beta_i]_n,[clsf_i]_n) = class \in [0,1]
\end{equation}
Throughout our investigation, we use six ($n = 6$) known domains gathered from the PhysioNet challenge and four diverse unseen databases (see Section \ref{dbs}).
 
\subsection{Data Selection\label{dbs}}

Table \ref{dbsall} outlines the selected databases for the experiment. As mentioned, we use the sub-databases of the PhysioNet database as known domains. This selection is for two reasons. Firstly, the PhysioNet database is the most commonly used database for heart sound related studies (both classification and segmentation) \cite{Messner2018HeartNetworksb}. Secondly, this database is a collection of diverse sub-databases that we can use to demonstrate the common scenarios that might occur during domain generalization. As unseen domains, we use the Fetal Heart sound database \cite{Samieinasab2015FetalSeparation}  which we split into groups containing all the mothers and all the fetuses, and the M3dicine databases collected by ourselves \cite{Fernando2019HeartAttention}. As mentioned in Table \ref{dbsall}, these unseen databases were collected from various devices, with different  experimental settings and with different  patient types including fetuses and animals. Therefore, by adopting these domains as unseen domains, we believe that our investigation will provide an extensive overview of how to accurately classify such diverse domains using domain generalization. Furthermore, since there is no direct link between the M3dicine-Animal database or the Fetus database and the selected known domains, outcomes from domain-recognition will help to understand what particular domains contain similar patterns, while not mapping direct relationships.  

The second and third columns of the table show the number of normal and abnormal instances in the datasets. These instances are computed using a 1s window of the signal with an appropriate window shift (for PhysioNet and M3dicine: 0.2s, Fetal databases: 0.5s). To eliminate noise at the start and at the end of the signal, we remove a portion of the signal which is chosen according to the database size and the duration of the wave file (on average we remove 0.5s from the beginning and at the end). For training the classifiers, we use the Mel Frequency Cepstral Coefficients (MFCCs) of the 1s signal. To compute this spectrum, we use 26 Mel-Filter Banks within the frequency range 0-600Hz. The resulting feature map has the shape $[26\times99\times1]$.

\subsection{One-Shot Learning Model\label{tripsec}}

In their FaceNet paper, Schroff and Philbin \cite{Schroff2015FaceNet:Clustering} introduced Triplet-Loss for face recognition tasks in One-Shot Learning. The primary objective of their investigation was to determine a unique embedding for a given image in the source domain, which then can be used to determine the similarities and differences between new instances. Ideally, this embedding should have similar values even if the pose of the face changes. Building on their strategy, we use the Triplet-Loss to find a unique embedding (or a signature) for a domain that is domain-relevant. To our knowledge, we are the first study that extends this strategy to domain generalization with class variability. 

\textbf{Triplet Loss}: As the name implies, a triplet contains three samples namely: the anchor, negative and positive. In the context of our problem, we define,

\begin{enumerate}
    \item \textbf{Anchor~(a)}: instance sampled from a domain.
    \item \textbf{Positive~(p)}: instance belonging to the same domain (may belong to the same or different class).
    \item \textbf{Negative~(n)}: instance belonging to a different domain.
\end{enumerate}

The triplet loss is given in Equation \ref{tripl}, where the $margin$ is defined as the separation between the anchor and the negative (default to 1.0) and d is the Euclidean (L2) distance between the normalized embeddings,
\begin{equation}
\label{tripl}
    \mathcal{L}_{tr} = max(d(a,p)-d(a,n) + margin,0).
\end{equation}
Considering all possible combinations, we can identify three types of triplets. 

\begin{enumerate}
    \item \textbf{Easy Triplets}: Triplets which have a zero loss, $d(a,p) + margin > d(a,n)$.
    \item \textbf{Hard Triplets}: In these triplets, the negative is closer to the anchor than the positive. In the context of our problem, the positive might be in a different class compared to the anchor.  
    \item \textbf{Semi-Hard Triplets:} 
    The negative is not closer to the anchor, but still has a positive loss: $d(a,p) < d(a,n) < d(a,p) + margin$
\end{enumerate}

The process of identifying these triplet types is called Triplet mining. According to \cite{Schroff2015FaceNet:Clustering}, the online mining of Semi-Hard Triplets tends to yield a steady learning process which will ultimately provide a better-optimized solution. Therefore, in this investigation, we use a similar approach. As discussed, the Triplet method yields a unique embedding for image classification tasks. Here, in this research, we use this method for identifying domains with two classes (normal/abnormal). This will add additional complexity to the model because now the model has to separate domains that have class differences within them. Therefore, to allow the model to see the differences between the classes, we train the model using a combined loss function. 

\begin{figure*}[h!]
    \centering
    \includegraphics[width=\linewidth]{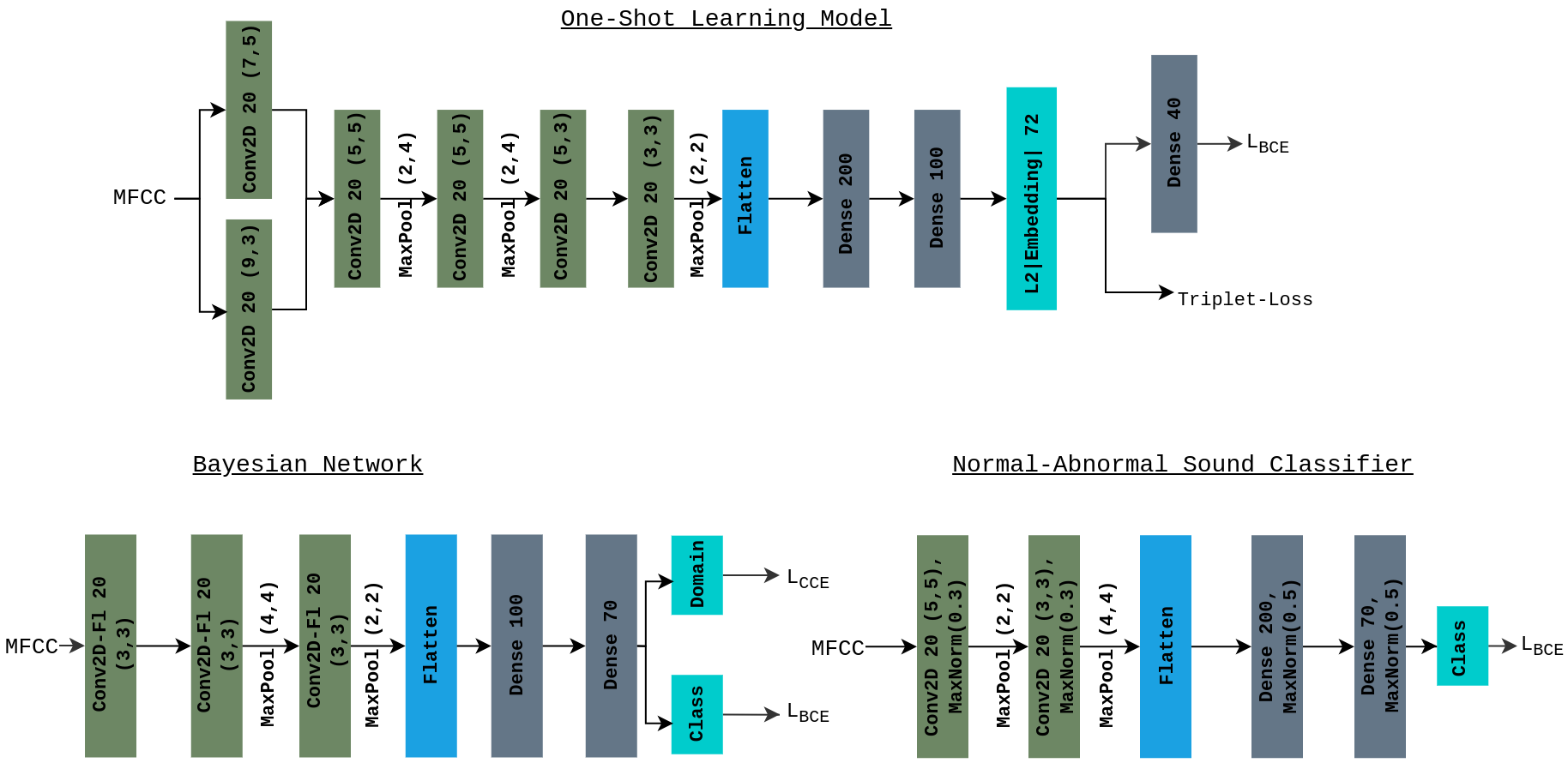}
    \caption{Network architectures proposed in the investigation. \textbf{Top}: The One-Shot learning model proposed in \ref{tripsec}, \textbf{Bottom-Left}: The architecture of the Bayesian Network proposed in \ref{bcnnsec}, \textbf{Bottom-Right}: The architecture of the normal/abnormal classifier in \ref{probsec}. \textit{Conv2D-Fl XX (h,w)}: Convolution 2D layers using Flipouts with \# kernels, hight, width of kernels. \textit{Dense-Fl XX}: Fully connected dense layers using Flipouts  with \# of neurons, $L_{BCE}$: Binary Cross-Entropy loss, $L_{CCE}$: Categorical Cross Entropy loss.}
    \label{siamese}
\end{figure*}

The $\top$ diagram of Figure \ref{siamese} displays the proposed network architecture. The input to the model is an MFCC feature map of a 1s signal window computed using 26 Mel-Filter Banks. The deep learning model consists of six 2D convolution layers (Conv2D) and two fully connected (Dense) layers. The model has two outputs: the embedding is shaped $[72\times1]$  (L2 normalized), and a classification output is obtained by extending the embedding using a dense layer with 40 neurons, 
\begin{equation}
\label{tripl-bros}
\mathcal{L} = \theta\times\mathcal{L}_{tr} + \alpha\times\frac{1}{N} \sum_{i=0}^{N} (y\times\log(\hat{y}_i) - (1-y)\times\log(1-\hat{y}_{i})).
\end{equation}
Equation \ref{tripl-bros} shows the combined loss function for two outputs of the proposed model. As expressed in Equation \ref{tripl-bros}, for the binary classification task, we use binary cross entropy ($L_{BCE}$) ($y$ is the true class label, $\hat{y}$ is the predicted class, $N$ is the number of instances).

The output embedding of the model is a unique signature associated with a particular domain. In the test phase, the embedding generated by the model for an unseen domain is used to find the domain-instance relationship. In the context of the generated embedding, we define the association as the number of similar instances in a given domain to the unseen instance ($I_{us}$). This similarity measure is computed using the L2 distance (i.e the Euclidean distance). If samples are dissimilar, then the L2 norm should be greater than the margin (in this case 1.0). For a given domain $D_i$ with $N$ samples, we compute the similarity measure by finding the number of instances that falls under the defined L2 normalized threshold ($\lambda, 0.0 \leq \lambda \leq 1.0$),
\begin{equation}
\label{threshod}
    \beta_i = \frac{n}{N},~ n = count([d(I_{us},I_j) < \lambda, ~\forall ~I_j~\in ~D_i]).
\end{equation}
After computing the relationship factors for each known domain, the instance is classified using an ensemble of classifiers. See Section \ref{probsec} for details regarding the final classifier. 

\subsection{Multi-Task Bayesian Model\label{bcnnsec}}

The second strategy we used to identify the domain-instance relationship is inspired by Bayesian Neural Networks. In a typical scenario, an instance from an unseen domain can hold a One-to-One or One-to-N relationship(s) with selected basis domains. If the relationship is One-to-One, then directly developing a domain classification model will solve the problem effectively. However, the developed model will not be able to capture relationships such as One-to-N, which can be seen as a typical scenario in medical data. Therefore, acknowledging this fact, we employ a Bayesian learning approach to develop a classification model that can recognize the most suitable domain(s) given an unseen instance. 

Equation \ref{bnneq1} describes the theoretical function used to determine the Bayesian-Inferred prediction of the model using the training data $D$, inputs $x$, outputs $y$ and the model weights (parameters) $\theta$.

\begin{equation}
    p(y|x,D) = \int_{\theta} p(y|x,\theta_i) p(\theta_i|D) d\theta_i.
    \label{bnneq1}
\end{equation}

In practice, the predictive probability and uncertainty is approximated using a Monte-carlo approach. Using this technique, we sample weights ($\theta_{i}$) from the variational distribution ($q_\phi(\theta)$), and compute the model's predictions conditioned on the sampled weights. Simply put, we are predicting the output for a given input using an ensemble of deep learning models ($NN_{\theta_i}(x)$). Equation \ref{bnneq2} presents the complete procedure to compute the output. Here, N is the number of Monte-carlo simulations (N=100),

\begin{equation}
    \hat{y} = \sum_{i=1}^{N} NN_{\theta_i}(x). 
    \label{bnneq2}
\end{equation}

After computing predictive probabilities using sampled weights ($\theta_i$), the final prediction of the model is taken as the average of the predictions made by all models (we which we term the predictive probability). Furthermore, intuitively, the uncertainty is quantified as the variance of the prediction, and here in this analysis, we ignore the uncertainty of the prediction.

\textbf{Flipouts}: Methods introducing stochasticity to deep learning models and algorithms can be categorized into two types: network-weight perturbing techniques and network-activation perturbing techniques. One commonly used method for network-activation perturbation is the Dropout technique introduced in \cite{Srivastava2014Dropout:Overfitting}. Among methods that employ network-weights  perturbation, the DropConnet approach introduced by Wan~et~al.~\cite{pmlr-v28-wan13} randomly disables (i.e sets to zero) the weights in the network to make partially activated neurons instead of probabilistically disabling neurons as in Dropout. 

Examining these two types of techniques, weight perturbations are computationally expensive because the neural network holds a larger number of weights compared to the number of activation units. Considering the implementation details of such methods, activation-based techniques use the entire mini-batch, which ultimately decreases the variance of the stochastic gradients by 1/N (N is the mini-batch size). In contrast, most of weight regularization methods utilize a single sample from the mini-batch. Wen~et~al.~\cite{Wen2018FLIPOUT:MINI-BATCHES} identify this as the major limitation of weight-perturbing methods, and introduce the Flipout technique that can achieve variance reduction in deep networks by computing weight-perturbations for each sample (example) in the mini-batch. 

Bayesian networks can also be identified as a network-weight perturbation technique where the network attaches a probability distribution to its parameters (weights and biases). The Flipout method can be also applied to Bayesian networks, and it has the ability to offer fast convergence to the learning process. Therefore, acknowledging this fact, we use Flipout as an additional regularization technique. Furthermore, since the domain classification should be based on the normal/abnormal heart sound recordings within each domain, we use a multi-task learning strategy. By adding this additional classification task, the model will have the ability to learn the domain difference while being able to understand the differences between classes \cite{Bi2008AnDiagnosis,Zhang2017ALearning}. 
 
As mentioned, the ultimate goal of stage one of the algorithm is to identify the relationship between the unseen instances and the known domains. Using this Bayesian inference technique, we directly represent the relationship as a probabilistic prediction. For the domain classification task, we use the categorical cross-entropy ($L_{CCE}$) loss function (Equation \ref{cat}). Here, $y$ is the true class, $\hat{y}$ is the predicted class, $N$ is the number of instances in the training set and $M$ is the number of classes (or in this case domains, $M$=6),

\begin{equation}
\label{cat}
\mathcal{L}_{CCE}(y,\hat{y}) = \sum_{j=0}^{M}\sum_{i=0}^{N}(y_{ij}\times \log(y_{ij})). 
\end{equation}

For the classification task (normal or abnormal), we use the binary cross entropy loss function (Equation \ref{bin}), 

\begin{equation}
\label{bin}
\mathcal{L}_{BCE}(y,\hat{y}) = \frac{1}{N} \sum_{i=0}^{N} (y\times\log(\hat{y}_i) - (1-y)\times\log(1-\hat{y}_{i})).
\end{equation}

\begin{equation}
\label{combloss}
\mathcal{L} = \theta\times\mathcal{L}_{CCE} + \alpha\times\mathcal{L}_{BCE}. 
\end{equation}

The combined loss function ($\mathcal{L}$) is in the form of Equation \ref{combloss} where $\theta$ and $\alpha$ are determined experimentally.

The proposed Bayesian network has a similar architecture compared to our previous work (abnormal heart sound classification using PhysioNet) \cite{dissanayake2020understanding}. After conducting an evaluation, we found that a simpler version of the same model can be used as a domain classifier. Given that a Bayesian network utilizes twice the number of parameters compared to a regular network with the same architecture, we reduced the number of convolution filters and the number of neurons in respective dense layers in our previous study to design the proposed Bayesian model. The bottom-left diagram in Figure \ref{siamese} shows the architecture of the proposed Bayesian Network. The network has three convolution layers and two dense layers. All of these layers use the Flipout technique to effectively drive the learning process. 

\subsection{Multiple Classifier Fusion\label{probsec}}

Strategies discussed in Subsections \ref{tripsec} and \ref{bcnnsec} implement two different deep learning techniques to solve the task of domain classification. After computing the source-target domain relationship values, we move on to the classification of the unseen domains using an ensemble of classifiers. This ensemble consists of models trained on each known domain (six in total). Since some of the adopted basis domains hold fewer examples, we use a slightly different architecture from our previous study \cite{dissanayake2020understanding}, which is the state-of-the-art model trained on the entire PhysioNet data. The adopted model in this study has two convolution layers instead of three, and each of these layers has 20 filters (compared to 60 and 40 in \cite{dissanayake2020understanding}). These models are trained on a balanced dataset since each database holds imbalanced data. The bottom-right diagram of Figure \ref{siamese} demonstrates the architecture of the classifier. 

As expressed in Equation \ref{probinf}, our framework takes the instance-domain relationship vector and the ensemble of classifiers trained on basis domains to compute the final outcome (i.e the given instance is normal or abnormal). Both domain recognition methods compute relationship vectors that indicate the association of the given unseen instance to basis domains. Among those relationship types, the same fusion process can be used to classify an instance with One-to-One and One-to-N relations. To be more precise, One-to-One can be interpreted as a subset of a One-to-N relationship. For such outcomes, we use majority voting. 

We compute the thresholded domain relationship vector using $thr()$ function. Here, $thr(v,\phi) = [1~\textbf{if}~i~>=~\phi ~\textbf{else}~0~~\textbf{for}~i~\in~~v]$, $v$ is a vector and $\phi$ is a hyperparameter. After computing the threshold, we obtain the majority class using class labels given by the models trained on related domains (i.e domains that are having 1 in the threshold vector). 
The output of the $thr()$ function will contain ones and zeros depending on the relationship type. For example, if the given instance represents a One-to-N relationship, then the resulting vector will contain multiple ones rather than being a one-hot vector that belongs to a One-to-One relationship. 

In the performance evaluation process, we use the $\lambda$ parameter in Equation \ref{threshod} to tune the outcome of the One-Shot learning based prediction. However, for the domain recognition output of the Bayesian model we adopt a different strategy. Since the model directly provides us with a probabilistic prediction ($p$), we use the $\phi$ parameter in the $thr()$ function to vary the output. By doing this, we are trying to find the optimal probabilistic threshold that will provide us with the best outcome. For instance, if we use a higher probabilistic threshold value ($p>0.7$), then we will end up with a smaller number of basis domains that show higher confidence values for the association. 

\begin{figure*}[t!]
    \includegraphics[width=\linewidth]{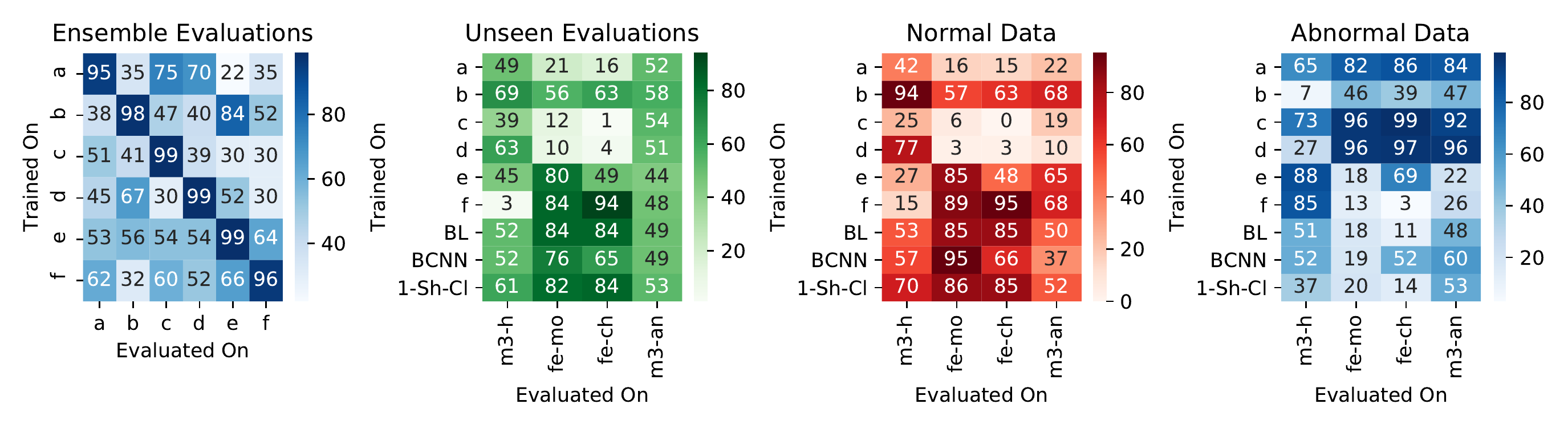}
    
    \hspace{1.8cm} (a) \hspace{4cm} (b) \hspace{4cm} (c) \hspace{4cm} (d)
    
    \caption{\textbf{(a)}~Ensemble model evaluation results, \textbf{(b)}~ Domain-accuracy relationship matrix, \textbf{(c)}~Evaluation results on normal samples, \textbf{(d)}~Evaluation results on abnormal samples.   
    Known: \textbf{PhysioNet~[a, b, c, d, e, f]}, Unseen: M3dicine-Human \textbf{m3-h}, Fetal-Mother's \textbf{fe-mo}, Fetal-Child's \textbf{fe-ch} and M3dicine-Animal \textbf{m3-an}, \textbf{BL}: Baseline model trained on the entire PhysioNet data, \textbf{BCNN}: Model introduced in  Section \ref{probsec}, \textbf{1-Sh-Cl}: Model introduced in section \ref{tripsec}}
    \label{dom-acc}
\end{figure*}

More importantly, by varying these hyperparameters, we will be able to effectively assess the performance of the fusion technique (for the One-Shot model $\lambda$ and for the Bayesian model $p$, see Table \ref{fres}). Finally, we may encounter instances with One-to-None relationships. For these scenarios, we rely on the predictions made by the classification output of the domain recognition network. 

\subsection{Baseline Model.\label{secbl}}

As mentioned, there is no prior research on generalization approaches designed for biosignal data. Therefore, Therefore,  in selecting a baseline for our model, we adopt the study by \cite{Mancini2018BestNets} from the Computer Vision domain. The model proposed in Manciniet et al. \cite{Mancini2018BestNets} is a joint learning framework where the domain identifier and individual domain-specific classifiers are learned through a combined loss function.

The main difference between our study and their study is that we use a two stage learning approach whereas they use a combined learning approach. We further note that they have used the Rotated-MNIST dataset i.e a rotated version of the commonly used MNIST for their study by representing a domain-shift as a rotation effect on an image. As in this investigation we are dealing with complex domain shifted signal data in the audio domain with inter subject variability, there is a large difference in the data used in \cite{Mancini2018BestNets} and this work.

As in our study, the proposed architecture in \cite{Mancini2018BestNets} also has two major components: domain identifier (DI) and domain-specific classifiers (DSCs). Given that we already designed optimal architectures for DI and DSC components in our study, to ensure fairer comparison we use these same components for the baseline. More specifically, we use the bottom-right architecture in Figure \ref{siamese} as the DSC component, and bottom-left architecture in Figure \ref{siamese} as the domain identifier. Here, to change the architecture to a domain classifier, we removed the classification output, and also we changed the model to a traditional CNN network (i.e without Bayesian Learning).

The final prediction is computed using Equation \ref{baselineeq}, where $x$ is the input instance and $\hat{y}$ is the prediction made by the model. The model contains N classifier branches for domain specific learning ($f^{dsc}_i(x)$), which returns a prediction vector. The domain classifier ($f^{di}_i(x)$) returns a same shaped vector indicating the probabilities of a given sample belonging to each domain (softmaxed, sums to 1.0),

\begin{equation}
    \hat{y} = \sum_{i=1}^{N} f_{i}^{di}(x)~f_{i}^{dsc}(x)
    \label{baselineeq}
\end{equation}

Similar to the Bayesian model, the framework returns a domain classification output and a normal-abnormal classification output. Therefore, we use a similar loss function a training approach to train the baseline model. Furthermore, as in the Bayesian approach, we save the best domain classification architecture.

\section{Results\label{results}}

Our results are presented across five major sections. In the first section, we discuss the domain-accuracy relationship between known domains and unseen domains. This will provide insights into the underlying relationships between selected domains. Following that, in the second section, we explain the findings from the One-Shot learning model, and we visualize the resulting embeddings in $\mathcal{R}^2$ using t-SNE \cite{tsne}. In the third section, we examine the probabilistic predictions made by the Bayesian network. Then, in the fourth section, we show how to use these outcomes to accurately classify an unseen/new domain using an ensemble of deep learning classifiers, and we also evaluate individual performance gains of our multi-task domain recognizes. In this section, we compare our final performance to the baseline model discussed in Section \ref{secbl} which is based on \cite{Mancini2018BestNets}. In the final section we summarize our findings and discuss some of the limitations observed. It should be noted that, for all performance evaluations, we use the 10 Fold Cross-Validation strategy, and we train our models using the Adam optimizer for 300 epochs.

\begin{figure*}[b!]
\begin{subfigure}{.33\textwidth}
  \centering
  \includegraphics[width=.79\linewidth]{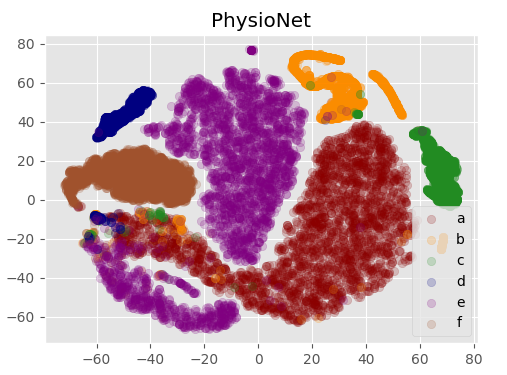}
  \caption{}
  \label{en}
\end{subfigure}
\begin{subfigure}{.33\textwidth}
  \centering
  \includegraphics[width=.79\linewidth]{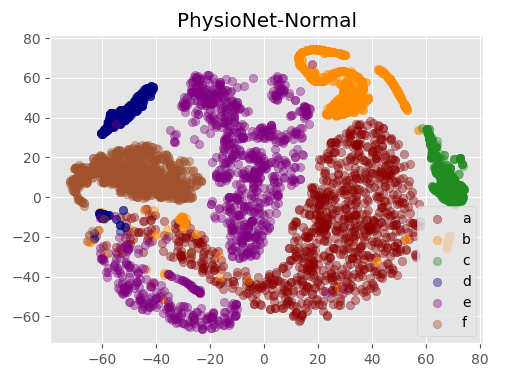}
  \caption{}
  \label{nor}
\end{subfigure}
\begin{subfigure}{.33\textwidth}
  \centering
  \includegraphics[width=.79\linewidth]{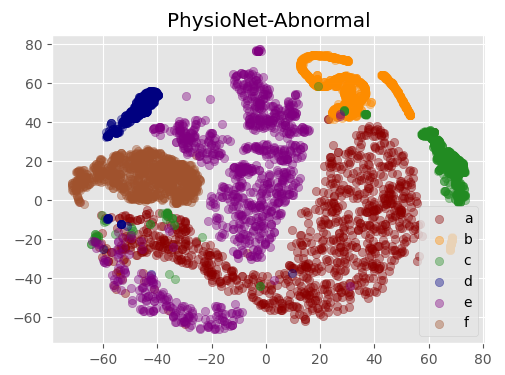}
  \caption{}
  \label{abn}
\end{subfigure}
\caption{t-SNE results for PhysioNet domains for 2000 randomly picked instances (normal: 1000, abnormal: 1000). \ref{en}: Visualization for the entire dataset, \ref{nor}: Normal samples, \ref{abn} Abnormal samples in the same distribution.}
\label{pca-nan}
\end{figure*}

\begin{figure*}[b!]
    \centering
    \includegraphics[width=0.99\linewidth]{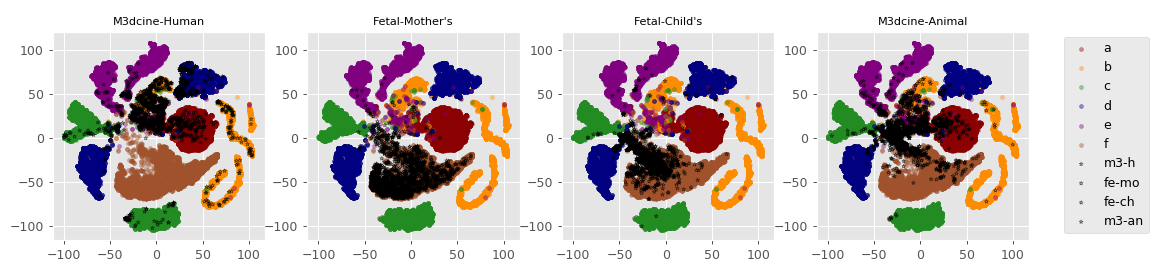}
    \caption{t-SNE results for unseen domains (black color *, 2000 random instances from each domain) with the PhysioNet data. \textbf{m3-h}: M3dicine Human, \textbf{fe-m}: Fetal-Mother's, \textbf{fe-c}: Fetal-Child's,  \textbf{m3-an}: M3dicine-Animal}
    \label{pca-us}
\end{figure*}

\subsection{Overview of Domain-Accuracy relationship}

Figure \ref{dom-acc} demonstrates average validation accuracy of each known domain and the domain-accuracy relationships between known domains and unseen domains. As mentioned, these known domain-based models are trained on a balanced database, and each of those models individually achieves almost 98\% validation accuracy after training (see the left plot in Figure \ref{dom-acc}). Furthermore, since some of the unseen databases adopted are highly imbalanced (see Table \ref{dbsall}), we also show the class-specific classification accuracies. 

According to the data shown in Figure \ref{dom-acc}, models trained on databases b and d perform well on M3dicine-Human data compared to the rest of the known domains. But the accuracy is limited to 60\%. In contrast, the Fetal databases seem to have a direct relationship to domain f. The Fetal-Mother's database achieves 84\% accuracy, which is expected as metadata from f indicates that the database has recordings from pregnant women. Furthermore, the database also contains recordings from athletes and patients with different medical conditions (asthma, tachycardia, bradycardia etc.), which make it more diverse compared to other databases. Furthermore, observing the domain-accuracy relationship between unseen M3dicine-Animal domain and the PhysioNet data, it is apparent that almost all domains tend show validation accuracies around 50\%.  

Given the imbalanced nature of the datasets used, it is important to consider class specific accuracies. Examining M3dicine-Human data, a model trained on basis domain b provides with the highest accuracy gain. In contrast, abnormal instances of the M3dicine-Human data demonstrates comparatively good accuracy on models trained on domains e and f.
Since this database has a comparatively high number of normal samples (see Table~\ref{dbsall}), the combined accuracy on domain b shows a relatively high value compared to other known domains. For normal data in Fetal databases, domain f performs well. For abnormal data, classifiers trained on domains c and d demonstrate good results. Even with these accuracy differences, still, domain f can be recognized as the most relevant domain because of the highly imbalanced nature of Fetal databases. For databases with relatively balanced data (eg: M3dicine-Human and Animal),  the basis classifiers perform differently depending on the class label.  However, for Fetal databases, this may not be a valid consideration as the database is highly imbalanced. 

These observed relationships are an outcome of various properties of the data. One obvious reason is the considered domains having similar recordings related to the same medical conditions observed in the unseen domain. Similarly, another reason might be the diversity within the known domains which helps to develop more generalized models.  

Collectively, it is apparent that most of the unseen domains have similarities with the known domains. However, compared to M3dicine-Human data, Fetal databases seem to associate well with the known domains (i.e shows good validation accuracies). 

\subsection{Results from the One-Shot Learning Model\label{resonsh}}

The resulting model achieves 0.34($\pm$0.01) training triplet loss and 0.22($\pm$0.01) validation triplet loss, and obtains a normal/abnormal sound classification accuracy of 81.20($\pm$0.23)\% ($\theta = \alpha = 1.0$). 
Since the model yields an embedding sized 72 ($\mathcal{R}^{72}$), we use t-SNE to understand how the model separates the domains.  It should be noted that the embedding size 72 is a result of a series of experiments carried out to determine the optimal dimension. As shown in Figure \ref{pca-nan}, even in two-dimensional space, we can observe a clear separation of domains. Furthermore, plots \ref{abn} and \ref{nor} illustrate the class distribution within each domain cluster.

\begin{figure*}[h!]
    \centering
    \includegraphics[width=0.95\linewidth]{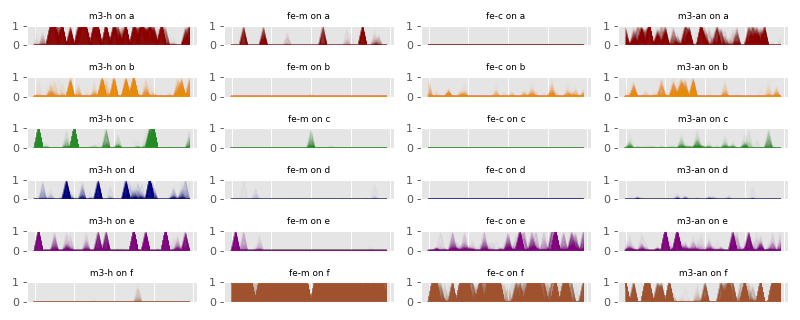}
    \caption{Predictions from the Bayesian Network for 100 continuous iterations indicating the known domains and unseen domains. The plots only illustrates 100 randomly picked normal and abnormal instances (50 each). \textbf{a,~b,~c,~d,~e,~f} from PhysioNet,  \textbf{m3-h}: M3dicine Human, \textbf{fe-m}: Fetal-Mother's, \textbf{fe-c}: Fetal-Child's, \textbf{m3-an}: M3dicine Animal}
    \label{bccn_prob}
\end{figure*}

After visualizing the domain separation for known domains, we apply t-SNE to unseen domains. Figure \ref{pca-us} illustrates how the unseen domain data is distributed among known domains. Here, we are trying to determine whether these domains possess domain-accuracy relationships observed in the previous section. To make the distribution more visible, we display samples of the unseen domains as black stars (*), and the plot only shows 2000 randomly picked data instances. According to t-SNE results, it is clear that the unknown domains hold the domain-accuracy relationship witnessed. 

Examining the embeddings for the M3dicine-Human data, the instances seem to be clustered among domains a, b and d which is reasonable when recognizing the accuracy relationship.  As expected, a majority of embeddings computed from the Fetal data fall to domain f compared to other domains. Furthermore, considering the M3dicine-Animal data, the embeddings present similarities to domains a, b, c and d. Collectively, along with the validation loss of the One-Shot learning model, this visualization demonstrates the model performs as expected. 

\subsection{Results from the Bayesian Network}

For the Bayesian Model, we implemented a multi-task classifier that can identify both domains and normal-abnormal classes. This classifier holds two hyperparameters that should be tuned (see Equation \ref{combloss}). After conducting a series of experiments and setting parameters $\theta$ and $\alpha$ to 1.0, we obtained a model with 93.12($\pm$0.19)\% and 86.17($\pm$0.16)\% validation accuracies for domain classification and normal/abnormal classification. It should be noted that after training the model we prioritise attention to the domain classification loss, and retained the best performing model using model checkpoints in the Keras API~\cite{chollet2015keras}.     

The resulting classifier has the ability to provide probabilistic predictions related to the instance domain relationship. Rather than relying on one single prediction, we determine the final probabilistic relationship through 100 continuous predictions (Monte-carlo simulations). 

Figure \ref{bccn_prob} illustrates the probabilistic predictions made by the classifier for each known domain (each row, PhysioNet~[a, b, c, d, e, f]). Each column of the figure represents an unseen domain starting from M3dicine-Human to M3dicine-Animal.

Examining the results, it is apparent that the Bayesian Network also captures the accuracy-domain relationship adequately. For M3dince-Human data, the probabilistic predictions indicate that the known domains a, b, c, d and e are related. However, we notice a higher confidence value for domain a compared to domain b which has the highest performance on the M3dicine-Human data. Aside from that outcome, the model seems to be able to capture the domain-accuracy relationship well on the Fetal databases. For both databases, the model presents great confidence for known domain f, which is expected as part of domain f holds recordings from pregnant women. However, the metadata does not indicate whether the domain possesses samples from fetuses or not. In additional to this similarity, both of the datasets  have used the same medical acquisition device (see Table \ref{dbsall}). For the M3dicine-Animal data, the model appears to identify a, b, e and f databases as related domains.

Collectively, it is apparent the Bayesian Model has the ability to capture the domain-accuracy relationship.

\begin{table*}[ht!]
    \centering
\begin{tabular}{|p{0.8cm}|p{0.9cm}|p{1.0cm}|p{1.0cm}|p{1.2cm}|p{1.2cm}|p{1.2cm}|p{1.0cm}|p{1.2cm}|p{1.2cm}|p{1.2cm}|p{1.0cm}|}
 \hline
 \multirow{2}{*}{DB} & \multirow{2}{*}{BL} & \multirow{2}{*}{BCNN} & \multirow{2}{*}{1-Sh-Cl} & \multicolumn{3}{|c|}{OneSh+Ense} & \multirow{2}{*}{Gain (\%)} & \multicolumn{3}{|c|}{BCNN+Ense} & \multirow{2}{*}{Gain (\%)} \\ \cline{5-7} \cline{9-11}
 
  & & & & $\lambda~<~0.2$ & $\lambda~<~0.5$ & $\lambda~<~1.0$ &  & $p>0.2$ & $p>0.5$ & $p>0.7$ & \\
  \hline
  m3-h & 52.27 & 52.12 & 61.03 &\cellcolor[gray]{.8} 68.45 & 67.33 & 66.70 & +16.18 & 61.04 & 58.73 & \cellcolor[gray]{.8} 62.17 & +9.90 \\
  fe-mo & 84.00 & 76.12 & 82.47  & \cellcolor[gray]{.8} 85.77 & 85.19 & 85.33 & +1.77 & 84.60 & 84.39 & \cellcolor[gray]{.8} 84.97 &  +0.97 \\
  fe-ch & 84.90 & 65.56 & 84.78  & \cellcolor[gray]{.8} 94.98 & 89.36 & 91.37 & +10.08 &  80.89 & 85.82 & \cellcolor[gray]{.8} 93.45 & +8.55 \\
  m3-an & 49.60 & 49.12 & 53.09 & 49.40 & 48.80 & \cellcolor[gray]{.8} 51.64 & +2.04 & 50.34 & 49.81 & \cellcolor[gray]{.8} 50.45 & +0.85 \\
 \hline\hline
 ph-net & 88.14 & 86.17 & 81.20 & 99.04 & 99.06 & 99.06 &  +10.92 & 99.10 & 98.05 & 98.98 & +10.96 \\ 
 \hline
\end{tabular}
    \caption{Final validation results and performance gains. \textbf{DB}: Database evaluated on \textbf{BL}: Baseline model by \cite{Mancini2018BestNets} discussed in Section \ref{secbl}, \textbf{BCNN}: Bayesian Convolution Neural Network from Section \ref{bcnnsec}, \textbf{1-Sh-Cl}: The extended model discussed in Section \ref{tripsec}, \textbf{OneSh+Ense}: One-Shot learning-based final model, \textbf{BCNN+Ense}: Bayesian Network-based final model, $\lambda$ and $p$ are hyperparameters of the respective models. The last column shows the validation results on PhysioNet (\textbf{ph-net}).}
    \label{fres}
\end{table*}

\subsection{The Complete System}

In the previous two sections, we discussed the results from the two learning techniques used to determine the instance domain relationship. Examining the combined results from both methods on unseen data  we see that: 
\begin{enumerate}
    \item \textbf{M3dicine-Human}: Both techniques indicate that the a,b,c,d,e domains have similarities to the M3dicine-Human data. The t-SNE visualization from the One-Shot learning model indicates b  has a strong relationship between M3dicine-Human data. In contrast, the Bayesian model shows higher confidence value for domain a. 
    \item \textbf{Fetal-Mother's}: Both techniques are showing similar results. 
   \item \textbf{Fetal-Child's}: Both techniques are showing similar results.
   \item \textbf{M3dicine-Animal}: Both techniques are showing similar results.
\end{enumerate}

We used the following evaluation protocol to determine how effectively the proposed system can be applied to achieve domain generalization. We trained the baseline model (BL) by adopting the learning strategy from \cite{Mancini2018BestNets} (discussed in Section \ref{secbl}). After conducting a series of evaluations, we found that setting $\theta$ = 0.9  and $\alpha$ = 0.1 yields the best performing model. Here,  as in our Bayesian model, we consider the domain classification accuracy as the measure of robustness of the model. The final model achieves 88.14($\pm$0.28)\% for normal-abnormal classification on the PhysioNet dataset and 89.29($\pm$0.27)\% for domain classification respectively.

In the evaluation process, we compare the baseline model's performance with two types of generalization models/frameworks. The first method is the proposed two stage classifier fusion system (named Fusion-based); the second is the multi-task learning strategy for the domain recognition neural networks, and we compare their performance on unseen domains (named Model-Based techniques).

\begin{itemize}
    \item \textbf{Fusion-based:} We demonstrate the final results from the One-Shot learning-based framework (OneSh+Ense) and the Bayesian Model-based framework (BCNN+Ense) on unseen domains.
    \item \textbf{Model-based:} Since we adopted a class-aware learning strategy for domain-recognizers, we also present the classification results from those models (BCNN: the multi-task Bayesian model, 1-Sh-Cl: Classification model using the extended embedding).
\end{itemize}
 
Table \ref{fres} summarizes the validation accuracies of the proposed deep learning models (cross-validated). The first column presents the unseen database used for evaluation. The second column of the table shows the validation results of the baseline model (BL). The next two columns present  classification accuracies of BCNN and 1-Sh-Cl models (i.e model-based techniques). The remaining eight columns depict the classification accuracies with the respective hyperparameters (here, $\lambda$ denotes the threshold value in Equation \ref{threshod}, $p$ is the probabilistic value predicted by the BCNN model and $\lambda,~p\in[0.0,~1.0]$) and individual performance gains. Furthermore, in the last row, we demonstrate the validation accuracies of the proposed models on the known PhysioNet domains. 

\textbf{Fusion-based:} According to the results presented in Table \ref{fres}, both proposed Fusion-based techniques  achieve higher performance for the classification tasks compared to the baseline model of \cite{Mancini2018BestNets}.  Examining results on specific unseen domains, both methods show high accuracy gains for the unseen human databases, and the accuracy for the M3dicine-Animal database is similar to the BL model. Furthermore, examining the classification performance of known (basis) domains, the final fusion method proposed also offers an approximately 10\% accuracy gain compared to the BL model. Therefore, collectively, our proposed learning techniques outperform the chosen baseline model  adopted from \cite{Mancini2018BestNets}, with significant classification accuracy gains on both known and unknown domains.

Considering the accuracy gains of the two proposed frameworks, for the M3dicine-Human data, the OneSh+Ense model achieves a higher accuracy gain compared to the BCNN+Ense model. Both classifiers produce almost the same results for the Fetal-Mother's database. However, the OneSh+Ense model performs better on Fetal-Child's instances compared to the BCNN+Ense model. This might due the One-Shot learning model identifying domain f as the most relevant domain for that particular database. However, compared to the previous unseen domains, both models have lower accuracy gains for the M3dicine-Animal data.

Examining all, the results show that the OneSh+Ense model outperforms the BCNN+Ense model. Since the M3dicine-Animal data consists of samples from animals and as we do not possess a matching database in the source domains, the method struggles. We believe the differences within the systole and diastole periods and frequency ranges primarily contribute to these differences. Furthermore, another important insight is that the accuracy increment seems to be the same or slightly less than the best individual performance gain (see Figure \ref{dom-acc}). However, this gain is significantly higher than the results from the baseline and multi-task BCNN model.

As stated, we used two hyperparameters to tune the outputs of the proposed models. Considering the OneSh+Ense model, $\lambda<0.2$ provides relatively high classification accuracies on unseen domains compared to other values. This indicates that estimating relationships with respect to the nearest instances produce the best outcome. Similarly, in the BCNN+Ense model, if the confidence of the probabilistic prediction is high for a given set of basis domains, then the fusion process provides relatively good results. However, collectively, changing the hyperparameters of the fusion models does not seem to produce significantly different results. 

\textbf{Model-based:} Examining accuracy gains of Model-based techniques (intermediate), for a majority of unseen databases, the 1-Sh-Cl model performs better than the baseline. The 1-Sh-Cl model demonstrates higher accuracy gain for the M3dicine-Human dataset compared to the baseline model,  and it also offers relatively good results for the M3dicine-Animal dataset. The same classifier achieves 84.78\% validation accuracy for Fetal-Child's data, which is similar to the accuracy gain of the baseline model. However, the baseline model offers slightly higher accuracy gain for the Fetal Mother's database. Examining these results, the 1-Sh-Cl model itself can be recognized as having greater generalisation ability compared to the baseline, and it shows uniform classification capability among all unseen domains (+2.65\% average gain). However, the validation accuracy on PhysioNet data shows lower results compared to the baseline.

In contrast, the BCNN model does not seem to offer higher classification gains compared to the baseline when considered as an individual classifier. For example, the model offers comparatively poor results for the Fetal databases. However, the model shows similar classification accuracies for M3dicine human and animal databases. For the basis domains, the model gives 86.17\% accuracy which is similar to the baseline's performance. However, for domain classification, the model shows significantly higher accuracy compared to the baseline. 

Examining the learning strategies adopted, the BCNN model and the BL model used a multi-task learning strategy, but the two paradigms are very different. 
In fact, the BCNN model can be seen as an ensemble of multi-task classifiers whereas the BL model is a complete classifier that fuses results from a non-varied ensemble. Despite the fact that they both demonstrate similar accuracies for the PhysioNet classification task, the BCNN model performs better for domain classification. We believe this contributes to superior results of the BCNN+Ense in all unseen domains.

Collectively, it is clearly noticeable that the 1-Sh-Cl model demonstrates comparatively good accuracies on all unseen domains compared to the baseline (average 2.65\% gain). Considering the BL model's substantial performance variation between Fetal and the M3dicine datasets, we conclude that the BL model does not show a uniform prediction capability across multiple unseen domains.

\subsection{Summary}

Regarding the performance of the baseline, it offered comparatively good accuracy for the Fetal databases, but performed poorly on the M3dicine data. Therefore, the model does not show a uniform prediction capability on unseen domains. This inferior generalisation performance  of the baseline system adopted from \cite{Mancini2018BestNets}  may result from the low domain classification accuracy of the model. We believe the combined training strategy proposed by \cite{Mancini2018BestNets} reduces the learning capacity of the DI and DSC components. Therefore, as demonstrated in our evaluation, training those components separately, as we have proposed, provides the optimal solution for the problem. 

Examining the known domain validation results, it is apparent that, compared to the baseline, the Fusion models have been able to sustain the classification performance on basis domains while being able to perform well on the unseen data. Therefore, this result validates the adopted strategy: dividing the problem into two stages does improve the performance compared to solving both simultaneously.

As noted, we evaluated the performance of our intermediate classification components on the unseen data. Examining all, the 1-Sh-Cl model shows good results for the task (average gain of 2.65\%), and it acts as a single classification entity without fusion. Therefore, if the learning method is powerful enough, even a single deep learner is able to perform adequately, and we believe 1-Sh-Cl to be the most computationally efficient model, and we encourage researchers to conduct further research. Compared to this, the BCNN model showed a poor generalization capability. 

The output of a fusion model depends on the ability of the domain-recognition model as a faulty prediction will lead to the instance being evaluated on an unrelated classifier(s), which can lead to an incorrect result. At this stage, dividing the domain generalization task into two separate stages does improve the results compared to training a single model for both tasks. Examining accuracy gains of the final model compared to the identified best performing domain, the final accuracy seems to be less than (or the same) compared to the highest observed accuracy gain. One of the reasons for this might be the lower diversity of the predictions made by the models in the ensemble. If those models classify uniformly (i.e the same set of instances are being correctly predicted), then, that scenario does not improve the classification accuracy significantly. 

As illustrated in Equation \ref{threshod}, to compute probabilistic values (or the domain classification), we utilize the entire training dataset. Given that the  basis domains have significant differences, and more importantly, they have inner variations, we were obliged to use the entire dataset. A sampling algorithm would be a possible approach, but the maximum sample size we can adopt is 1800, given that domain b contains 900 normal samples (from 3000 samples total). However, databases such as a and e contain more varied instances, and using a smaller sample may not clearly represent the entire distribution. Therefore, considering this limitation, the final prediction is made by comparing a new instance with the entire dataset, and here, we store the predicted-embeddings of the PhysioNet data with corresponding domains and the trained model. By doing this, we make our fusion process more efficient, and the data size needed to be stored for a classification process is comparatively small (instances shaped $[72,]$ vs $[26\times99\times1]$, float64 type). We acknowledge this as one of the limitations in our system, and we encourage researchers to conduct further investigations regarding modelling distributions of such varied datasets.

\section{Conclusion\label{discuss}}

The objective of this investigation was to formulate and  implement a deep learning system that can achieve generalization on biosignal classification tasks. Being the first study that investigates this problem in the medical field, we believe that the insights gained from this work will help researchers to understand how such a system should be implemented, and more importantly, what sort of factors should be taken into consideration while developing such systems. 

Compared to the static nature of images, biosignals possess temporal characteristics, which makes the domain separation process more difficult. The first learning method we introduced, the Triplet Loss, comes from computer vision classification tasks and we believe our implementation demonstrates that an extension of same method can be applied to domain recognition. Along with this technique, the Bayesian method also tends to offer competitive results for domain identification. Collectively, it is apparent that both introduced techniques have the ability to effectively address the domain separation task, and the resulting deep learning frameworks demonstrate classification accuracy gains up to 16\%.   

As discussed, the medical domain has diverse databases which encompass different medical equipments, patient types, acquisition protocols, etc. Having such diverse domains helps us to identify the differences within those domains, and assist in designing more generalized deep learning architectures for medical applications.


%



%

\bibliographystyle{IEEEtran}
\bibliography{references,bibadd}

%

\end{document}